\documentclass[runningheads]{llncs}

\usepackage{eccv}

\usepackage{eccvabbrv}

\usepackage{graphicx}
\usepackage{booktabs}
\usepackage{multirow}

\usepackage[accsupp]{axessibility}  % Improves PDF readability for those with disabilities.

\usepackage[pagebackref,breaklinks,colorlinks]{hyperref}
\usepackage{orcidlink}

\begin{document}

\title{Text Image Generation for Low-Resource Languages with Dual Translation Learning} 

\titlerunning{Abbreviated paper title}

\author{Chihiro Noguchi \and
Shun Fukuda \and
Shoichiro Mihara \and
Masao Yamanaka 
}

\authorrunning{C.Noguchi et al.}

\institute{Toyota Motor Corporation \\
\email{chihiro\_noguchi\_aa@mail.toyota.co.jp}
}

\maketitle

\begin{abstract}
Scene text recognition in low-resource languages frequently faces challenges due to the limited availability of training datasets derived from real-world scenes. This study proposes a novel approach that generates text images in low-resource languages by emulating the style of real text images from high-resource languages. Our approach utilizes a diffusion model that is conditioned on binary states: ``synthetic'' and ``real.'' The training of this model involves dual translation tasks, where it transforms plain text images into either synthetic or real text images, based on the binary states. This approach not only effectively differentiates between the two domains but also facilitates the model's explicit recognition of characters in the target language. Furthermore, to enhance the accuracy and variety of generated text images, we introduce two guidance techniques: Fidelity-Diversity Balancing Guidance and Fidelity Enhancement Guidance. Our experimental results demonstrate that the text images generated by our proposed framework can significantly improve the performance of scene text recognition models for low-resource languages.
\end{abstract}

\begin{figure}[h!]
    \begin{tabular}{c}
      \begin{minipage}[t]{0.95\linewidth}
        \centering
        \includegraphics[width=11.18cm]{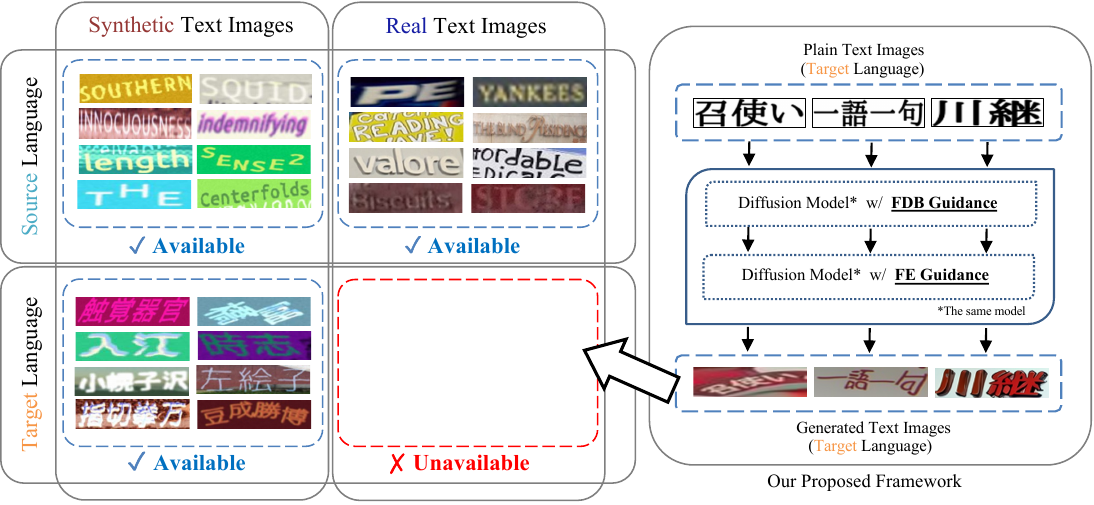}
      \end{minipage}
      
    \end{tabular}
    \caption{While the source language has both synthetic and real text images available, the target language only possesses synthetic text images. This study aims to generate text images in the target language that emulate the style of real text images.}
\label{fig:top_figure}
\end{figure}

\section{Introduction}
\label{sec:introduction}

Scene text recognition \cite{7801919,fang2021read,bautista2022scene,baek2021if,baek2019wrong} has attracted significant attention due to its high applicability in various areas. In recent years, numerous training datasets sourced from real scene images have become publicly available. These datasets facilitate the development of text recognition models that offer robust performance in real-world settings. However, the majority of these datasets focus on major languages, especially English and Chinese. Consequently, for low-resource languages, the prevalent strategy is to resort to synthetically created text images. These synthetic text images produced by existing methodologies inevitably exhibit a domain gap when compared to real text images. Such a domain gap arises primarily because of their restricted diversity. The creation of synthetic text images typically involves a limited selection of fonts, text effects, background visuals, and simple synthetic degradation techniques like Gaussian blur and noise. As a result, text recognition models trained on these synthetic images often struggle to maintain robust performance in real-world scenarios. This study aims to generate text images that bridge this domain gap and enhance scene text recognition performance for low-resource languages.

Many studies have focused on the task of translating images from one domain to another \cite{isola2017image,wang2018high,rombach2022high}. This line of methods are promising to make more realistic text images. Unsupervised image-to-image translation \cite{zhu2017unpaired,kim2017learning,yi2017dualgan} aims to learn the translation function between two image sets. Although this approach allows for training a model that translates from synthetic to real images, it might neglect the consistency of textual content before and after the translation, which can result in translated images either lacking or presenting altered textual content. Another line of studies involves style transfer \cite{karras2019style,abdal2019image2stylegan,deng2022stytr2,zhang2023inversion,wang2023stylediffusion}. This approach focuses on extracting styles from a real image and applying them to a synthetic one. Such techniques have been extended to text images \cite{krishnan2023textstylebrush,lyu2017auto,zhang2018separating,yang2019tet}, with an emphasis on preserving the textual content of the synthetic images. However, this approach basically assumes that text in all images belong to the same language.

In this study, we propose a framework to generate text images for low-resource languages, capturing the styles of real text images from a high-resource language, including realistic degradation and diverse text styles. As illustrated in Fig. \ref{fig:top_figure}, we explore a practical scenario where synthetic text images are accessible in both languages, but real text images are only available in the high-resource language. Henceforth, we refer to high-resource languages as source languages, and low-resource languages as target languages. For the generation of accurate text images suited for training recognition models, it must meet two essential criteria: (1) The framework must capture the styles of real text images from the source language, ensuring realistic degradation and diverse text styles, irrespective of the textual content. (2) It must comprehend the textual content of the target language. This enables the generation of text images that contain significantly deformed characters while preserving the original textual content.

To satisfy the first criterion, we utilize a diffusion model (DM) \cite{ho2020denoising,NEURIPS2021_49ad23d1,pmlr-v139-nichol21a,ho2022cascaded} conditioned on a binary variable with two states: \textit{synthetic} or \textit{real}. When set to \textit{synthetic}, the model is trained to produce synthetic text images, whereas when set to \textit{real}, it is trained to produce real images. This training approach is effective in differentiating between the two domains of text images, irrespective of the text's language. For the second criterion, we utilize plain text images, characterized by a white background and a single font, to condition the DM on the textual content. This setup leads to dual translation tasks: transforming plain text images into synthetic ones and into real ones. Incorporating training for the model to translate from plain to synthetic images is crucial, as it enables the DM to comprehend characters in the target language. We refer to this training strategy as Dual Translation Learning (DTL). It's important to note that the plain-real images solely include text in the source language, while the plain-synthetic images contain text from both the source and the target languages.

To generate more precise and diverse text images, we introduce two guidance techniques for inference. The use of classifier-free guidance \cite{ho2021classifierfree} has been recognized as an effective method to improve image quality. Our observations indicate that the guidance scale notably influences the balance between textual content fidelity and diversity in the generated text images. To achieve an ideal balance, we propose Fidelity-Diversity Balancing (FDB) Guidance. This approach schedules the guidance scales from lower to higher values to maintain both fidelity and diversity. In addition, we propose Fidelity Enhancement (FE) Guidance to further increase the fidelity to the textual content. To enhance fidelity while minimally affecting the styles derived from real text images, we utilize concepts from diffusion-based image translation \cite{meng2022sdedit,Choi_2021_ICCV,gao2023back,peng2023diffusion,kwon2023diffusionbased}. Collectively, these guidance strategies effectively enhance the fidelity and diversity in the generation of text images.
Our contributions are summarized as follows.
\begin{itemize}
    \item We introduce a novel framework for text image generation, featuring DTL guided by a binary state. This approach is particularly effective in emulating the style of real text images and in precisely comprehending and rendering characters in the target language.
    \item We introduce two guidance techniques: FDB and FE Guidance. These strategies greatly improve the precision and diversity of text image generation.
\end{itemize}

\section{Related Works}

\subsection{Synthetic Text Image Generation}

\noindent
\textbf{Text Rendering Engines.}\quad
MJ \cite{jaderberg2016synthetic} and ST \cite{gupta2016synthetic} are the most popular text image synthesis engines. MJ creates text images by employing a series of rendering modules, including font rendering, border/shadow rendering, base coloring, projective distortion, natural data blending, and noise injection. On the other hand, ST overlays multiple texts onto scene images, followed by cropping text boxes from these images. The resulting text images can contain text noises from other text boxes, simulating real-world scenarios. Zhan et al. \cite{zhan2018verisimilar} introduced synthesis methods designed to make text images conducive to accurate and robust scene text detection and recognition. Their approach uses semantic coherence, visual saliency, and an adaptive text appearance model. Recently, SynthTIGER \cite{yim2021synthtiger} was introduced to amalgamate the strengths of both MJ and ST approaches. It achieves comparable performance to the combined dataset of MJ and ST.

\noindent
\textbf{Style Transfer.}\quad
Utilizing image generative models such as GANs \cite{goodfellow2014generative} provides an alternative method for generating text images. Given two types of images---style images and content images---the goal of style transfer is to extract solely the style features from the style images and apply them to the content images, while preserving the original textual content.
Text editing \cite{nakamura2017scene,zhang2019ensnet} aims to replace the textual content of a scene image with alternative content. SRNet \cite{10.1145/3343031.3350929} and SwapText \cite{yang2020swaptext} are seminal text editing frameworks, encompassing modules for text conversion, erasing, and fusion. STEFANN \cite{roy2020stefann} particularly targets character-level regions to facilitate precise replacement. The training of these models necessitates paired text images, which are commonly sourced from synthetic text images. TextStyleBrush \cite{krishnan2023textstylebrush} and RewriteNet \cite{lee2021rewritenet} utilize pretrained text recognition models to facilitate alignment between the target textual content and the generated text image.  While the use of pretrained text recognition models can enhance the clarity and accuracy of text images, they tend to produce simpler samples for training datasets. Consequently, text images that the pretrained model struggles to recognize are not generated. 
Font style transfer \cite{lyu2017auto,zhang2018separating,yang2019tet} represents an another line of research, focusing on the exploration of font styles with the goal of generating text images that reflect the font used in style images.
Although many studies within this field operate under a few-shot paradigm \cite{azadi2018multi,park2021multiple,tang2022few,li2021few,liu2022xmp,wang2023cf,fu2023neural}, where a few style images are provided, such a scenario does not align with our objectives. This deviation arises as collecting real images with a coherent image style is too expensive.
Recently, DG-Font \cite{xie2021dg,chen2022dgfont++} has been proposed, achieving one-shot font style transfer, presenting a viable methodology for our intended purpose.

\noindent
\textbf{Image-to-image Translation.}\quad
Image-to-image translation methods \cite{isola2017image,wang2018high,rombach2022high} can be used to translate synthetic text images into realistic ones.
However, this approach necessitates paired images for training, making them unsuitable for the current setting. In contrast, unsupervised image-to-image translation, which allows for the translation of an image from one domain to another without requiring paired images, is applicable. Such methods often enforce cycle consistency \cite{zhu2017unpaired,kim2017learning,yi2017dualgan} across the two domains. DA-GAN \cite{zhan2019ga} and SF-GAN \cite{zhan2019spatial} further refine this by breaking down cycle consistency into spatial and appearance components, facilitating the generation of text images that accommodate domain shifts in both spaces.

\subsection{Diffusion Models}

The concept of Gaussian DMs was initially presented in \cite{pmlr-v37-sohl-dickstein15} and has since undergone significant enhancements in the context of image generation \cite{ho2020denoising,NEURIPS2021_49ad23d1,pmlr-v139-nichol21a,ho2022cascaded,zhu2023conditional}. Owing to their training stability, these models are versatile enough to be conditioned on data from a range of modalities, including images \cite{hu2023gaia}, text \cite{rombach2022high,Noguchi_2024_WACV}, and audio \cite{shen2023difftalk}. At the inference phase, the application of classifier guidance \cite{ho2021classifierfree} has emerged as an essential technique for enhancing output quality. Recently, diffusion-based image-to-image translation has attracted significant attention. This method can be used without paired images, making it versatile for a wide range of applications \cite{Choi_2021_ICCV,gao2023back,peng2023diffusion,kwon2023diffusionbased}. Furthermore, DMs have found application in creating text images \cite{liu2022character,yang2024glyphcontrol,tuo2023anytext,chen2024textdiffuser}. However, their goals differ from ours; they primarily seek to enhance the ability of Stable Diffusion \cite{rombach2022high} to produce unblurred and readable text images. Consequently, they rely on pretrained OCR models. In contrast, our goal is to enhance the OCR models themselves.

\section{Methodology}

\subsection{Preliminaries: Diffusion Models}
Our proposed framework is based on denoising diffusion probabilistic models (DDPMs). In this subsection, we briefly revisit the DDPM.

Diffusion models consist of forward and reverse processes.
Given a data distribution ${x}_0\sim q({x_0})$, the forward process is defined as the Markov process, where a series of latent variables $x_1,\dots,x_T$ is produced by progressively adding Gaussian noise $q(x_t|x_{t-1}) = \mathcal{N}(\sqrt{1-\beta_t}x_{t-1},\beta_t\boldsymbol{I})$
to the sample. Hence, $q(x_{1:T}|x_0)=\prod_{t=1}^Tq(x_t|x_{t-1})$.
Here, $\beta_t\in(0,1)$ indicates the variance at time $t$.
When $T$ is sufficiently large, $x_T$ is equivalent to a pure Gaussian noise.
Due to the reproductive property of Gaussian, we can directly sample $x_t$ at any time step $t$ from a single Gaussian with $x_0$ input as follows:

\begin{equation}
    q(x_t|x_0) = \mathcal{N}(\sqrt{\Bar{\alpha}_t}x_0, (1-\Bar{\alpha}_t)\boldsymbol{I}),
    \label{eq:q_xt_x0}
\end{equation}
where $\alpha_t=1-\beta_t$ and $\Bar{\alpha}_t=\prod_{s=0}^t\alpha_t$.

In the reverse process, starting from a Gaussian noise $x_T\sim\mathcal{N}(0,\boldsymbol{I})$, we can reverse the forward process by sampling from the posterior $q(x_{t-1}|x_t)$.
This posterior $q(x_{t-1}|x_t)$ can be approximated as a Gaussian distribution when $\beta_t$ is sufficiently small \cite{pmlr-v37-sohl-dickstein15}.
Therefore, $q(x_{t-1}|x_t)$ can reasonably fit to the true posterior by being parameterized as $p_\theta(x_{t-1}|x_t) = \mathcal{N}(\mu_\theta(x_t,t),\Sigma_\theta(x_t,t))$.

The loss function is provided by the variational lower bound of the negative log likelihood $\mathcal{L}=\mathbb{E}[-\log p_\theta(x_0)]$.
The covariance $\Sigma_\theta$ is set as a constant in many cases, whereas the mean $\mu_\theta$ is parameterized via a deep neural network using a UNet \cite{ronneberger2015u} architecture. The problem of estimating the mean $\mu_\theta$ can be reformulated equivalently as the one of estimating Gaussian noise $\epsilon$ contained in $x_t$. As a result, the final loss function is obtained as follows:

\begin{equation}
    \mathcal{L}=\mathbb{E}_{t,x_0,\epsilon}[\|\epsilon-\epsilon_\theta(x_t,t)\|^2].
    \label{eq:unconditional_ddpm_loss_app}
\end{equation}
See \cite{pmlr-v37-sohl-dickstein15,ho2020denoising} for the detailed derivation of Eq. \ref{eq:unconditional_ddpm_loss_app}.

To control the generation of images, a conditional DM $p_\theta (x_0|y)$ with the condition $y$ is commonly used.
According to prior studies, this can be easily achieved by simply conditioning the denoising model $\epsilon_{\theta}$.

\begin{figure*}[t!]
    \begin{tabular}{c}
      \begin{minipage}[t]{0.95\linewidth}
        \centering
        \includegraphics[width=12cm]{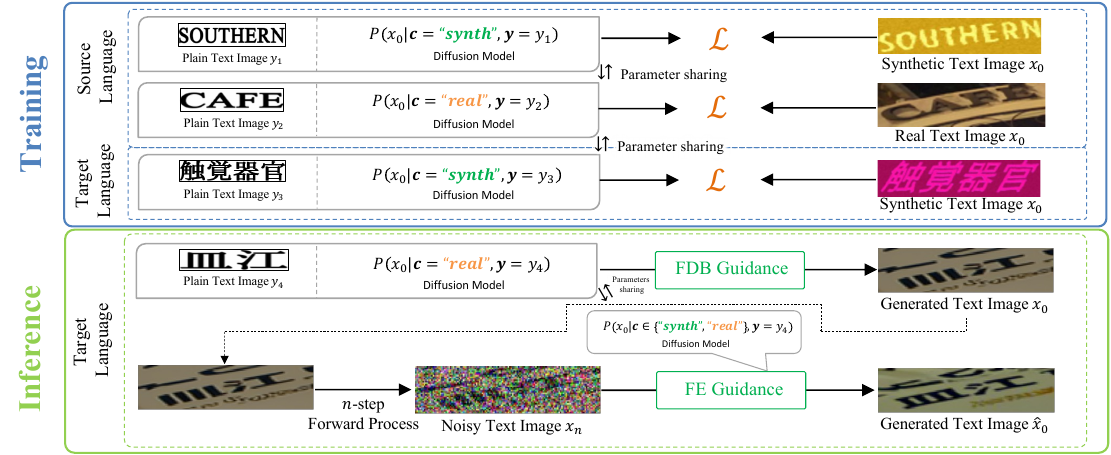}
      \end{minipage}
      
    \end{tabular}
    \caption{Overview of the proposed framework for text image generation. At the training phase, the DM is trained to generate synthetic and real text images, governed by a binary state input:  either \textit{synth} or \textit{real}. Plain text images are used as input to provide their corresponding textual content. At the inference phase, plain text images in the target language are fed into the model under the \textit{real} condition. FDB Guidance empowers the model to generate text images with enhanced precision and variety. Moreover, FE Guidance can further improve the text content fidelity of the generated text images.}
\label{fig:overview}
\end{figure*}

\subsection{Problem Setup}
This study addresses two distinct domain shifts.
The first arises between source and target languages, where the source language is defined as one having ample availability of real text images and their corresponding text labels, whereas the target language lacks such datasets.
The second domain shift comes from the difference between synthetic and real text images; real text images are obtained from real-world scenes, while synthetic text images are created using a text rendering engine.
The main goal of this study is to generate text images in a target language while applying the style of real text images in a source language.
In our proposed framework, we utilize a rendering engine to produce two kinds of text images: plain and synthetic.
Plain text images are created with a single font against a white background.
In contrast, synthetic text images are created with various fonts, backgrounds, and additional synthetic styles such as text colors, textures, effects, geometric transformations, and degradations.
We represent text images in the source language with $\{\mathcal{S}^{plain}, \mathcal{S}^{synth}, \mathcal{S}^{real}\}$ and those in the target language with $\{\mathcal{T}^{plain}, \mathcal{T}^{synth}\}$. Here, $\mathcal{S}^{plain}$ and $\mathcal{T}^{plain}$ denote plain text images, $\mathcal{S}^{synth}$ and $\mathcal{T}^{synth}$ signify synthetic images, and $\mathcal{S}^{real}$ stands for real images.

\subsection{Text Image Generation Framework}
\label{sec:text_image_generation_framework}

\noindent
\textbf{Dual Translation Learning.}\quad
The overview of the proposed framework is illustrated in Fig \ref{fig:overview}.
In this framework, the DM is conditioned on two types of inputs. The first is plain images, which serve to supply textual content. The second is a binary variable, denoted as $c$, with two states:  
\textit{synth} and \textit{real}. 
At the training phase, when $c=synth$, the DM is trained to generate synthetic images.
In contrast, when $c=real$, the DM is trained to generate real images. 
Here, we denote the conditional DM as $p_\theta (x_0|c,y)$, where $y$ stands for a plain image corresponding $x_0$.
When $c=synth$, $p_\theta (x_0|c,y)$ is trained with $x_0\in\{\mathcal{S}^{synth},\mathcal{T}^{synth}\}$ and its corresponding $y\in\{\mathcal{S}^{plain},\mathcal{T}^{plain}\}$.
When $c=real$, it is trained with $x_0\in\mathcal{S}^{real}$ and its corresponding $y\in\mathcal{S}^{plain}$.

At the inference time, we set $c=real$ and give $y\in\mathcal{T}^{plain}$ to the DM. Although the conditional model $p(x_0|c=real,y)$ is not trained on text images from the target language, it is capable of generating text images in that language, emulating the styles of real text images. 
This ability arises because the DM can learn to recognize characters of the target language through the translation task from $y\in\mathcal{T}^{plain}$ to its corresponding $x_0\in\mathcal{T}^{synth}$. Without this training, it generates corrupted text images when $y\in\mathcal{T}^{plain}$ is inputted. Moreover, by training the DM under the two conditions, $c=real$ and $synth$, using text images in the same source language, it can discern the style difference between the synthetic and real text images.

\begin{figure}[t!]
    \begin{tabular}{c}
      \begin{minipage}[t]{0.95\linewidth}
        \centering
        \includegraphics[width=11.5cm]{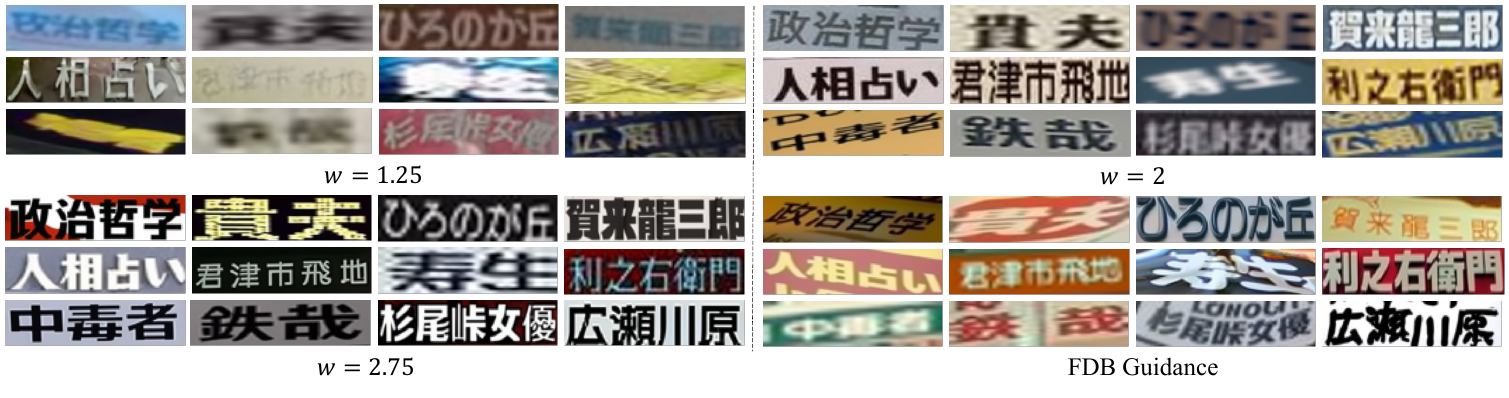}
      \end{minipage}
      
    \end{tabular}
    \caption{Examples of text images generated using a constant guidance scale $w\in \{1.25, 2, 2.75\}$, and using FDB Guidance.}
\label{fig:samples_gs}
\end{figure}

\noindent
\textbf{Fidelity-Diversity Balancing Guidance.}\quad
The DM in our proposed framework adopts classifier-free guidance \cite{ho2021classifierfree} to generate text images that more accurately reflect the textual content of the input plain text images. The classifier-free guidance is formulated as
\begin{equation}
    \tilde{\epsilon}_{\theta}(x_t, t, c, y)=\epsilon_{\theta}(x_t, t, c, y)+w(\epsilon_{\theta}(x_t, t, c, y)-\epsilon_{\theta}(x_t, t)),
    \label{eq:classfier_free_guidance}
\end{equation}
where $w$ denotes a guidance scale.
At the training time, this approach probabilistically omits the conditioning $c$ and $y$ at a consistent rate, leading to a joint model for both unconditional and conditional objectives.

Based on our experiments, the guidance scale has a marked impact on the fidelity and diversity of the generated text images. With a smaller guidance scale, the resulting images display a variety of text styles, fonts, and noise patterns (high diversity). However, the textual content of these images often deviates from their corresponding plain images (low fidelity). Conversely, with a larger guidance scale, the textual content of the generated images aligns more closely with the plain images, but at the expense of reduced diversity. In Fig. \ref{fig:samples_gs}, we showcase text images generated using different guidance scales.

To harmonize fidelity with diversity, we schedule the guidance scale in relation to $t$. Specifically, we initiate with a minimal guidance scale and progressively increase it in a linear fashion. That is, $w_t = (t/T)w_{min} + (1-(t/T))w_{max}$, where $w_{min}$ and $w_{max}$ stand for the minimal and maximal guidance scales, respectively. In the early stage of the reverse process (when $t$ is large), the DM focuses on forming the overarching structure of the image. In this phase, using a small guidance scale can facilitate the generation of a diverse range of text images.
As the reverse process progresses (with $t$ decreasing), the model shifts its attention to refining the detailed structure of the image. Therefore, by increasing the guidance scale as $t$ decreases, we can enhance the model's ability to rectify the inconsistencies and inaccuracies in generated images relative to the intended textual content.

\noindent
\textbf{Fidelity Enhancement Guidance.}\quad
Although the FDB Guidance is effective to generate high-quality text images, the DM still occasionally generates text images misaligned with the input textual content. To rectify this, our framework introduces an additional guidance mechanism. 

As the conditional model $p(x_0|c=synth,y)$ is trained directly on text images from the target language, it can generate text images with higher fidelity. However, these text images inevitably exhibit the styles of synthetic text images. To achieve the fidelity enhancement without compromising the styles of real text images, we utilize techniques from diffusion-based image-to-image translation \cite{Choi_2021_ICCV,meng2022sdedit,gao2023back,peng2023diffusion,kwon2023diffusionbased}. For a text image $x_0$, generated through the FDB Guidance, we apply the forward process up to a time step $t=n$ (where $0<n<T$), producing a noisy version of $x_0$, denoted as ${x}_n$. Subsequently, the reverse process is applied to $x_n$ to obtain $\hat{x}_0$. During this reverse process, we set both $c=synth $ and $real$ depending on the time step $t$. Specifically, we apply these two conditions in a $k_1:k_2$ ratio. For instance, when $(k_1, k_2)=(1, 6)$, we set $c=synth$ when $t$ modulo 7 is 0, and $c=real$ otherwise. By carefully selecting $n$, the forward process can sufficiently obscure the text details in $x_0$ without eradicating its textual content and the style. As a result, the subsequent reverse process can rectify the text in $x_0$ while preserving its original textual content and style. 
It is important to note that selecting $c=synth$ for all time steps in the reverse process would reduce the realistic noise and blur present in $x_0$. As a result, it is advisable to minimize the frequency of the $c=synth$ condition as much as possible.

\begin{figure*}[t!]
    \begin{tabular}{c}
      \begin{minipage}[t]{0.95\linewidth}
        \centering
        \includegraphics[width=8.5cm]{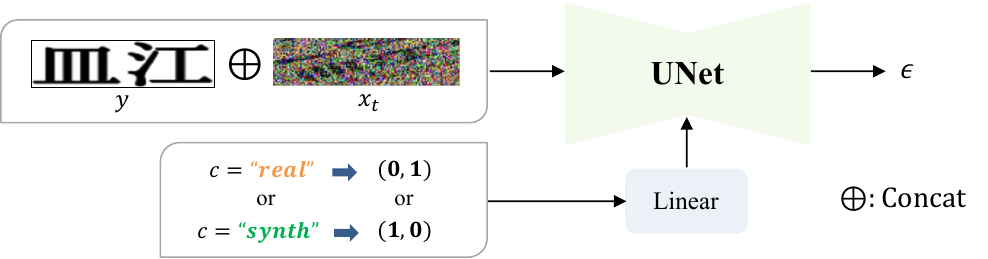}
      \end{minipage}
      
    \end{tabular}
    \caption{Diffusion model architecture in the proposed framework. Plain images are conditioned by being concatenated with input noisy images. Additionally, binary variables are conditioned through both timestep embeddings and cross-attention layers.}
\label{fig:model_architecture}
\end{figure*}

\noindent
\textbf{Diffusion Model Architecture.}\quad
Figure \ref{fig:model_architecture} presents an overview of the DM in the proposed framework. The framework utilizes the widely adopted UNet architecture similar to prior studies. To condition the DM on a plain image $y$, this image is concatenated with the input noisy image $x_t$. Furthermore, the DM is also conditioned on a binary variable $c$, which is incorporated into the diffusion timestep embedding. Additionally, the DM conditions on $c$ through the integration of cross-attention layers at multiple resolutions.

\section{Experiments}
To evaluate the effectiveness of the generated text images, we used off-the-shelf text recognition models trained with these images. Their recognition accuracy was used for the comparative metrics.

\noindent
\textbf{Dataset.}\quad
To train the DM in our framework, we require real text images in the source language, along with synthetic and plain text images in both the source and target languages.
We sourced the real text images from datasets for scene text recognition.
Due to the prevalent availability of English datasets, we selected English as the source language. 
We utilized ten real datasets: SVT \cite{6126402}, IIIT \cite{Mishra2009SceneTR}, IC13 \cite{6628859}, IC15 \cite{7333942}, RCTW \cite{shi2017icdar2017}, Uber \cite{zhang2017uber}, ArT \cite{chng2019icdar2019}, LSVT \cite{sun2019icdar}, MLT19 \cite{nayef2019icdar2019}, and ReCTS \cite{zhang2019icdar}.
Furthermore, we incorporated two extensive real datasets derived from Open Images \cite{krasin2017openimages}: TextOCR \cite{singh2021textocr} and annotations from the OpenVINO toolkit \cite{krylov2021open}.
These datasets comprise a total of 2.56M text images.
To produce the synthetic and plain text images, we utilized a recently proposed text rendering engine, SynthTIGER \cite{yim2021synthtiger}.

For the target languages, we selected five languages: Arabic, Bengali, Chinese, Japanese, and Korean. 
We utilized the training split of the MLT19 dataset, which contains text images in these languages, for evaluation. Using SynthTIGER, we produced 2M synthetic and plain text images for each language. The default settings of SynthTIGER were employed for this production, and we sourced the font files for each language from Google Fonts. The word set for each language was derived from the EasyOCR repository \cite{easyocr} and Wikipedia pages using an API. 
The training datasets consisted of 36 Arabic, 74 Bengali, 6,614 Chinese, 2,100 Japanese, and 1,471 Korean unique characters, respectively. Text images containing characters not included in these sets were excluded from the evaluation. Additionally, images of vertical text, characterized by greater vertical than horizontal length, were also excluded from the evaluation.
Detailed information about the datasets can be found in the Appendix.

\noindent
\textbf{Implementation Details.}\quad
For the training process, we employed the AdamW \cite{kingma2014adam} optimizer with a learning rate $1\times 10^{-4}$. Our DM underwent training for 1.5M iterations using a batch size of 128. During training, the diffusion time step $T$ was set to 1000. At inference, $T$ was set to 100 for the FDB Guidance and 200 for the FE Guidance. For the hyperparameters of these guidances, we selected $w_{min}=1$, $w_{max}=6$, and $n=120$. In addition, for $k_{1}$ and $k_{2}$, we set $(k_1, k_2)=(2,1)$ for Arabic, $(3,1)$ for Bengali, $(1,6)$ for Chinese, $(1,6)$ for Japanese, and $(1,1)$ for Korean. While a distinct DM was trained for each target language, all models utilized the same hyperparameters, with the exception of those related to the guidances.
All images, both for training and inference, were resized to a resolution of $32\times 128$. Further hyperparameter details can be found in the Appendix.

% \begin{table}[t!]
%   \centering
%   \resizebox{0.48\textwidth}{!}{
%   \centering
%   \begin{tabular}{c|ccccc}
%     \hline
%      & Arabic & Bengali & Chinese & Japanese & Korean  \\
%     \hline
%   $(k_1, k_2)$ & (2,1) & (3,1) & (1,4) & (1,6) & (1,1)\\
%     \hline
%   \end{tabular}
%   }
% \caption{$k_{1}$ and $k_{2}$ used for FE Guidance.}
% \label{table:guidacen_scales}
% \end{table}

\noindent
\textbf{Baselines.}\quad
To evaluate the quality of text images generated by our proposed framework, we used five baseline methods for comparison: (1) CycleGAN \cite{zhu2017unpaired}, (2) DG-Font \cite{xie2021dg}, (3) SynthTIGER \cite{yim2021synthtiger}, (4) SynthTIGER+Real-ESRGAN \cite{wang2021real}, and (5) SDEdit \cite{meng2022sdedit}.
(1) For the training of CycleGAN, we used 2M synthetic text images in both English and the target language, as well as the real text images. At inference, the trained model received synthetic text images produced by SynthTIGER.
(2) To train DG-Font, we used the real text images in English as style images and their corresponding plain text images as content images. At inference, plain text images from the source language served as content images, while real text images in English, chosen randomly, were employed as style images.
(3) The synthetic text images utilized in our experiments were created using SynthTIGER. For the evaluation of SynthTIGER's performance, we used the identical ones that were used for training the other baseline methods.
(4) Real-ESRGAN combines various synthetic degradation methods to simulate real-world degradations. We applied this pipeline with a 50\% probability at each training iteration to the text images created with SynthTIGER.
(5) SDEdit offers the most straightforward method for applying diffusion models to style transfer. Given synthetic text images, a forward process was executed up to a time step $t=n$ $(0<n<T)$. Subsequently, a reverse process was applied to the resulting noised images. This reverse process utilizes a diffusion model that was trained on real images, allowing it to transform synthetic images into ones that appear real. In this context, we evaluated two different DMs for use in the reverse process. One model was trained exclusively on real English text images. In contrast, the other model was trained with the DTL, employing the same DM with the condition $c=real$, as suggested by our method. We used the same value of $n$ as that used in the FE Guidance.

\noindent
\textbf{Evaluation Details.}\quad
In every experiment presented in this paper, 1M text images were generated using each of the baseline and our proposed methods. These text images were then used to train a text recognition model, PARSeq \cite{bautista2022scene}. The quality of the generated images was assessed through word accuracy (Acc.) and normalized edit distance (NED), which were derived using the PARSeq trained on these images.
The training of PARSeq adhered to its default configuration. For data augmentation, RandAugment \cite{cubuk2020randaugment} was employed, which offers 16 different transformations, such as Gaussian blur and Poisson noise, selecting three at random for every iteration.

\begin{table*}[t!]
  \centering
  \resizebox{0.95\textwidth}{!}{
  \begin{tabular}{c|cc|cc|cc|cc|cc}
    \hline
    \multirow{3}{*}{Method} & \multicolumn{10}{c}{Language}  \\
    % \cline{2-6}
    & \multicolumn{2}{c}{Arabic} & \multicolumn{2}{c}{Bengali} & \multicolumn{2}{c}{Chinese} & \multicolumn{2}{c}{Japanese} & \multicolumn{2}{c}{Korean} \\
    \cline{2-11}
     & Acc. & 1-NED & Acc. & 1-NED & Acc. & 1-NED & Acc. & 1-NED & Acc. & 1-NED \\
     \hline
  DG-Font \cite{xie2021dg} & 9.52 & 54.16 & 6.17 & 40.78 & 5.83 & 23.47 & 17.44 & 45.31 & 17.71 & 45.98  \\
  SynthTIGER \cite{yim2021synthtiger}  & 66.86 & 90.37 & 72.09 & 90.54 & 65.07 & 80.05 & 58.80 & 79.43 & 80.20 & 90.76  \\
  SynthTIGER + Real-ESRGAN \cite{wang2021real}  & 66.08 & 89.72 & 71.07 & 90.01 & 62.31 & 78.30 & 54.72 & 75.05 & 80.11 & 91.05 \\
  CycleGAN \cite{zhu2017unpaired} & 65.02 & 89.73 & 70.88 & 90.24 & 63.73 & 78.21 & 58.77 & 80.08 & 78.32 & 89.09\\
  SDEdit (w/o DTL) \cite{meng2022sdedit} & 46.45 & 81.96 & 58.68 & 83.85 & 49.20 & 68.69 & 54.56 & 74.89 & 69.73 & 85.67 \\
  \hline
  SDEdit (w/ DTL) & 66.83 & 90.37 & 69.55 & 89.60 & 68.36 & 82.50 & 62.63 & 82.86 & 80.34 & 91.19 \\
  Ours & \textbf{68.28} & \textbf{90.76} & \textbf{72.79} & \textbf{90.97} & \textbf{69.69} & \textbf{83.49} & \textbf{64.93} & \textbf{83.76} & \textbf{82.39} & \textbf{92.04} \\
    \hline
  \end{tabular}
}
\caption{Quantitative comparison with the existing text image generation methods. The last two methods were trained with dual translation learning (DTL).}
\label{table:comparison_with_sota_acc}
\end{table*}

\subsection{Comparison with Existing Methods}
\label{sec:comparison_with_existing_methods}
Table \ref{table:comparison_with_sota_acc} presents a comparison of our framework with existing methods. We can see that our proposed framework outperforms other methods across all languages. Notably, the performance improvement for Chinese and Japanese is considerably higher than for other languages. 
In addition, the results of SDEdit with DTL significantly surpass those achieved without DTL. Without DTL, the DM can mimic the style of real images yet struggles to accurately render text in the desired language. On the other hand, by incorporating DTL, the DM is able to generate accurate text images while still capturing the style of real images.
Additionally, we conducted a comparison with the case using Real-ESRGAN to assess the performance of simple combinations of synthetic degradation. While RandAugment offers basic degradation methods, Real-ESRGAN includes a broader and more intense range of degradation techniques.
Despite this, the results obtained with Real-ESRGAN were inferior to those achieved without it. These indicate that relying solely on synthetic degradation presents limitations in reproducing real-world degradation.

\begin{figure*}[t!]
    \hspace{-10pt}
    \begin{tabular}{c}
      \begin{minipage}[t]{0.95\linewidth}
        \centering
        \includegraphics[width=12.3cm]{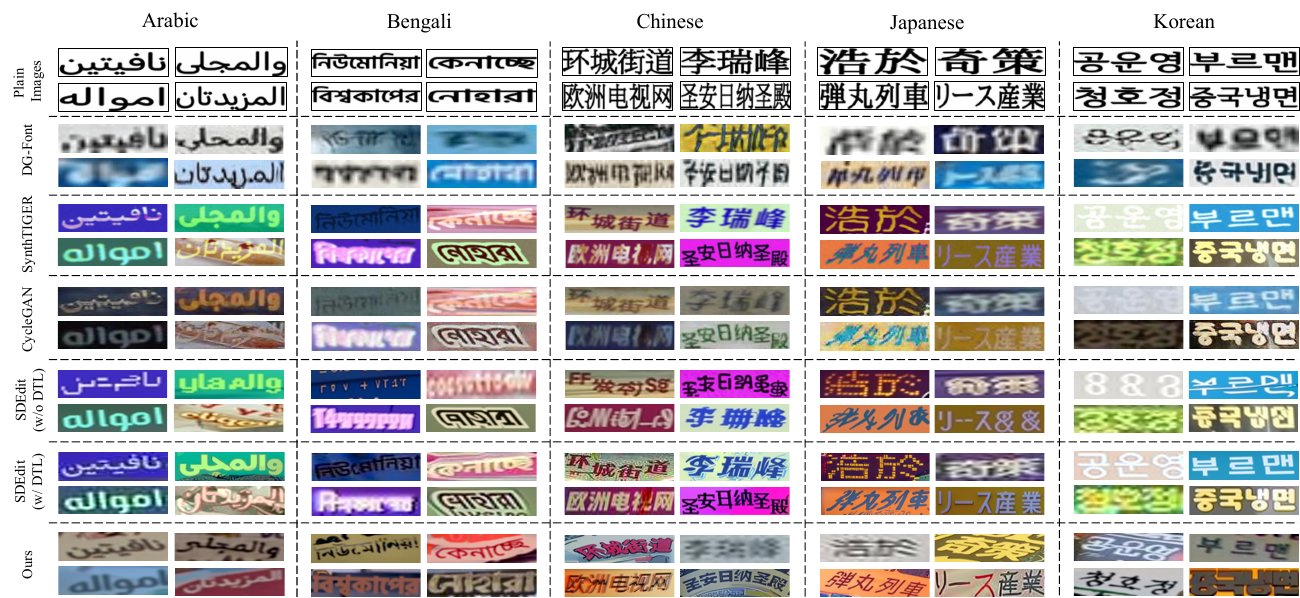}
      \end{minipage}
      
    \end{tabular}
    \caption{Qualitative comparison of generated text images. The top row displays plain text images, while the subsequent rows show text images generated by different methods, all sharing the same textual content as their corresponding plain text images.}
\label{fig:samples_qualitative_eval}
\end{figure*}

\subsection{Qualitative Evaluation}

In Fig. \ref{fig:samples_qualitative_eval}, we showcase samples created by the baseline methods alongside those from our proposed framework. DG-Font, trained exclusively on the source language, assumes a minimal disparity between the style and content images. Consequently, while it capably mimics the styles of real text images, it does not effectively preserve the textual content. Although CycleGAN seems to successfully preserve textual content and transfer style, the text images it produces show limited variation from the input synthetic text images generated using SynthTIGER, thus constraining their diversity. 
SDEdit can transfer the style from synthetic to realistic images, yet, in the absence of DTL, it struggles to maintain textual content. With DTL, it adeptly produces text images that retain textual content fidelity. In our framework, text images are not derived from synthetic images but are instead created anew from noise images. This approach enables the production of text images in a wide range of realistic styles, unaffected by the limited diversity of synthetic images.

\subsection{Ablation Studies}

\begin{table}[!t]
  \centering
  \resizebox{0.93\textwidth}{!}{
  \begin{tabular}{c|cc|cc|cc|cc|cc}
    \hline
    \multirow{3}{*}{Method} & \multicolumn{10}{c}{Language}  \\
    % \cline{2-6}
    & \multicolumn{2}{c}{Arabic} & \multicolumn{2}{c}{Bengali} & \multicolumn{2}{c}{Chinese} & \multicolumn{2}{c}{Japanese} & \multicolumn{2}{c}{Korean} \\
    \cline{2-11}
    & Acc. & 1-NED & Acc. & 1-NED & Acc. & 1-NED & Acc. & 1-NED & Acc. & 1-NED \\
    \hline
  w/o FEG & 60.45 & 87.75 & 65.54 & 87.69 & 69.51 & 83.29 & 64.75 & 83.56 & 80.80 & 91.32 \\
  w/ FEG & \textbf{68.28} & \textbf{90.76} & \textbf{72.79} & \textbf{90.97} & \textbf{69.69} & \textbf{83.49} & \textbf{64.93} & \textbf{83.76} & \textbf{82.39} & \textbf{92.04} \\
    \hline
  \end{tabular}
  }
\caption{Comparison of results with and without the use of FE Guidance (FEG).}
\label{table:comparison_with_without_FEG}
\end{table}

\noindent
\textbf{Effectiveness of FE Guidance.}\quad
To evaluate the effectiveness of the FE Guidance, we compared results with its utilization to those without.
As shown in Tab. \ref{table:comparison_with_without_FEG}, the FE Guidance enhances the accuracy of the text recognition model.
In particular, Arabic, Bengali, and Korean languages exhibit significant improvements. Considering that these languages contain characters that pose challenges for the DM to depict as discussed in Sec. \ref{sec:discussion}, they benefit more distinctly from the FE Guidance compared to other languages.

\begin{table}[!t]
  \centering
  \hspace{-15pt}
  \resizebox{0.43\textwidth}{!}{
  \begin{minipage}[t]{0.45\linewidth}
  \begin{tabular}{c|ccc}
    \hline
    \multirow{2}{*}{Guidance Scale} & \multicolumn{3}{c}{Language}  \\
    % \cline{2-4}
    & Arabic & Japanese & Korean \\
    \hline
    1.25 & 58.01 & 63.59 & 78.81  \\
  2 & 57.57 & 64.22 & 79.89   \\
  2.75 & 55.21 & 62.57 & 79.85   \\
  FDB Guidance & \textbf{60.45} & \textbf{64.75} &  \textbf{80.80}  \\
    \hline
  \end{tabular}
  \caption{Dependence on guidance scales.}
  \label{table:dependence_on_gs}
  \end{minipage}
  }
  \hspace{17pt}
\resizebox{0.43\textwidth}{!}{
\begin{minipage}[t]{0.45\linewidth}
  \begin{tabular}{c|ccc}
    \hline
    \multirow{2}{*}{Method} & \multicolumn{3}{c}{Recognition Model}  \\
    % \cline{2-5}
    & PARSeq & ABINet & TRBA \\
    \hline
    CycleGAN & 58.77 & 57.25 & 53.68 \\
  SynthTIGER & 58.80 & 58.41 & 54.64   \\
  SDEdit (w/ DTL) & 62.63 & 61.08 & 57.35   \\
  Ours & \textbf{64.93} & \textbf{64.29} & \textbf{61.43}   \\
    \hline
  \end{tabular}
  \caption{Comparison of results using different text recognition model.}
  \label{table:different_recognition_models}
  \end{minipage}
  }
\end{table}

\noindent
\textbf{Dependence on Guidance Scales.}\quad
As discussed in Sec. \ref{sec:text_image_generation_framework}, the guidance scale controls the trade-off between fidelity and diversity. As shown in Tab. \ref{table:dependence_on_gs}, using excessively large ($w=2.75$) or small ($w=1.25$) guidance scales results in inferior performance due to a diminished diversity and fidelity, respectively. While a guidance scale of $w=2$ offers improved performance relative to these results for Japanese and Korean, it involves concessions in both fidelity and diversity. Conversely, using FDB Guidance enables us to attain high fidelity without sacrificing diversity, thereby surpassing the results associated with the static guidance scales.

\noindent
\textbf{Evaluation using Different Recognition Models.}\quad
In Tab. \ref{table:different_recognition_models}, we present comparative results evaluated by three text recognition models: PARSeq \cite{bautista2022scene}, ABINet \cite{fang2021read}, and TRBA \cite{baek2019wrong}.
The performance trends of ABINet and TRBA are comparable to those observed for PARSeq, with our proposed method demonstrating superior performance over the others.

\begin{figure}[t!]
    \begin{tabular}{c}
      \begin{minipage}[t]{0.95\linewidth}
        \centering
        \includegraphics[width=11.5cm]{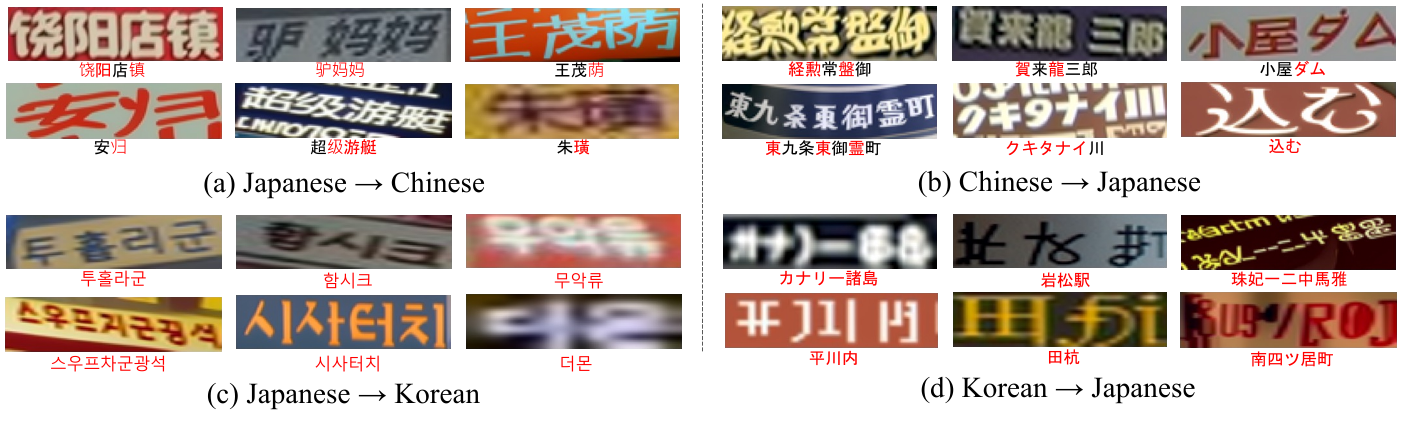}
      \end{minipage}
      
    \end{tabular}
    \caption{Examples generated by the proposed framework in the cross-language scenario. ``Language 1 $\to$ Language 2'' indicates that the DM was trained with text images from Language 1 as the target language, and subsequently, at inference, plain text images from Language 2 were used as input.}
\label{fig:samples_unseen}
\end{figure}

\begin{table}[!t]
  \centering
  \hspace{-30pt}
  \resizebox{0.4\textwidth}{!}{
  \begin{minipage}[t]{0.45\textwidth}
  \centering
  \begin{tabular}{c|ccc}
    \hline
    Target Language & \multicolumn{2}{c}{Language used for Inference}  \\
    at Training & Chinese & Japanese  \\
    \hline
    Chinese & \textbf{69.69} & 60.32 &  \\
  Japanese & 63.02 & \textbf{64.93} &  \\
    \hline
  \end{tabular}
\subcaption{}
\label{table:jp_vs_ch}
\end{minipage}
}
\hspace{30pt}
\resizebox{0.4\textwidth}{!}{
\begin{minipage}[t]{0.45\textwidth}
  \centering
  \begin{tabular}{c|ccc}
    \hline
    Target Language & \multicolumn{2}{c}{Language used for Inference}  \\
    at Training & Japanese & Korean  \\
    \hline
    Japanese & \textbf{69.69} & 70.86 &  \\
  Korean & 26.75 & \textbf{82.39} &    \\
    \hline
  \end{tabular}
\subcaption{}
\label{table:jp_vs_ko}
\end{minipage}
}
\caption{Comparison of results when the target language during training differs from the language of plain text images used as input at inference.  It showcases the comparative results between (a) Chinese and Japanese, and (b) Japanese and Korean.}
\end{table}

\subsection{Generalization for Unseen Characters}
Textual content is input into the DM through plain text images rather than through character identifiers. This implies that characters that were not defined during training could be processed at the time of inference. To investigate this generalizability, we first focused on the one across Chinese and Japanese text images. The resemblance of many characters between these languages potentially eases the cross-language generalization.
Figure \textcolor{red}{6a} presents samples of Chinese text images generated from the DM trained for Japanese, while Fig. \textcolor{red}{6b} illustrates the opposite case, showing images of Japanese text generated by a model trained for Chinese. Characters highlighted in red beneath each image denote those that were not included in the training set. The generated images successfully capture the intended textual content.
Additionally, we present the results of quantitative evaluation in Tab. \ref{table:jp_vs_ch}. While the results exhibit lower performance compared to scenarios with language consistency between training and inference phases, they still maintain substantial accuracy.

Figure \textcolor{red}{6c} shows samples of Japanese text images generated from the model trained for Korean, while Fig. \textcolor{red}{6d} presents the opposite scenario. The images in Fig. \textcolor{red}{6c} appear to successfully represent the intended textual content, while those in Fig. \textcolor{red}{6d} fail to do so. This discrepancy may be attributed to the extensive diversity of shapes within Japanese characters, which allows the model to synthesize unseen characters by recombining elements of known character shapes. Despite the unsuccessful depiction of the intended text in Fig. \textcolor{red}{6d}, there is an observable attempt to approximate Japanese text through the amalgamation of known Korean and English characters.

\section{Discussion and Limitations}
\label{sec:discussion}
Our empirical findings suggest that, for Chinese and Japanese text images, the DMs can achieve high fidelity to the intended textual content without relying on the $c=synth$ condition, thereby facilitating the generation of a wider variety of text images.
We posit that such text image diversity plays a pivotal role in the observed significant performance boost. In contrast, many characters in Arabic, Bengali, and Korean languages possess analogous shapes, making it challenging for the DM to precisely depict these characters without resorting to larger values of $k_1$. As a result, there is a constrained diversity in the generated text images for these languages, leading to a relatively marginal enhancement in performance. The pursuit of text image generation that maintains diversity across all target languages constitutes an avenue for future research.

\section{Conclusion}
This study presents a novel framework that effectively addresses domain gaps in text image generation for low-resource languages. Our proposed framework, which integrates binary-conditioned DMs trained with DTL, excels in not only emulating the styles of real text images but also comprehending textual content in target low-resource languages. The introduction of FDB and FE Guidance significantly improves the fidelity and diversity of the generated text images.
This study establishes a crucial foundation in the realm of text-image generation for low-resource languages, offering a significant step towards making text recognition technology more inclusive and universally accessible.

\clearpage  % TODO REVIEW/FINAL: This \clearpage needs to be removed from both review and camera-ready versions.

\bibliographystyle{splncs04}
\bibliography{egbib}

\begin{thebibliography}{10}
\providecommand{\url}[1]{\texttt{#1}}
\providecommand{\urlprefix}{URL }
\providecommand{\doi}[1]{https://doi.org/#1}

\bibitem{easyocr}
Easyocr. https://github.com/JaidedAI/EasyOCR

\bibitem{abdal2019image2stylegan}
Abdal, R., Qin, Y., Wonka, P.: Image2stylegan: How to embed images into the stylegan latent space? In: ICCV. pp. 4432--4441 (2019)

\bibitem{azadi2018multi}
Azadi, S., Fisher, M., Kim, V.G., Wang, Z., Shechtman, E., Darrell, T.: Multi-content gan for few-shot font style transfer. In: CVPR. pp. 7564--7573 (2018)

\bibitem{baek2019wrong}
Baek, J., Kim, G., Lee, J., Park, S., Han, D., Yun, S., Oh, S.J., Lee, H.: What is wrong with scene text recognition model comparisons? dataset and model analysis. In: ICCV. pp. 4715--4723 (2019)

\bibitem{baek2021if}
Baek, J., Matsui, Y., Aizawa, K.: What if we only use real datasets for scene text recognition? toward scene text recognition with fewer labels. In: CVPR. pp. 3113--3122 (2021)

\bibitem{bautista2022scene}
Bautista, D., Atienza, R.: Scene text recognition with permuted autoregressive sequence models. In: ECCV. pp. 178--196 (2022)

\bibitem{chen2024textdiffuser}
Chen, J., Huang, Y., Lv, T., Cui, L., Chen, Q., Wei, F.: Textdiffuser: Diffusion models as text painters. NIPS  \textbf{36} (2024)

\bibitem{chen2022dgfont++}
Chen, X., Xie, Y., Sun, L., Lu, Y.: Dgfont++: Robust deformable generative networks for unsupervised font generation. arXiv preprint arXiv:2212.14742  (2022)

\bibitem{chng2019icdar2019}
Chng, C.K., Liu, Y., Sun, Y., Ng, C.C., Luo, C., Ni, Z., Fang, C., Zhang, S., Han, J., Ding, E., et~al.: Icdar2019 robust reading challenge on arbitrary-shaped text-rrc-art. In: ICDAR. pp. 1571--1576 (2019)

\bibitem{Choi_2021_ICCV}
Choi, J., Kim, S., Jeong, Y., Gwon, Y., Yoon, S.: Ilvr: Conditioning method for denoising diffusion probabilistic models. In: ICCV. pp. 14367--14376 (2021)

\bibitem{cubuk2020randaugment}
Cubuk, E.D., Zoph, B., Shlens, J., Le, Q.V.: Randaugment: Practical automated data augmentation with a reduced search space. In: CVPR Workshops. pp. 702--703 (2020)

\bibitem{deng2022stytr2}
Deng, Y., Tang, F., Dong, W., Ma, C., Pan, X., Wang, L., Xu, C.: Stytr2: Image style transfer with transformers. In: CVPR. pp. 11326--11336 (2022)

\bibitem{NEURIPS2021_49ad23d1}
Dhariwal, P., Nichol, A.: Diffusion models beat gans on image synthesis. In: NIPS. vol.~34 (2021)

\bibitem{fang2021read}
Fang, S., Xie, H., Wang, Y., Mao, Z., Zhang, Y.: Read like humans: Autonomous, bidirectional and iterative language modeling for scene text recognition. In: CVPR. pp. 7098--7107 (2021)

\bibitem{fu2023neural}
Fu, B., He, J., Wang, J., Qiao, Y.: Neural transformation fields for arbitrary-styled font generation. In: CVPR. pp. 22438--22447 (2023)

\bibitem{gao2023back}
Gao, J., Zhang, J., Liu, X., Darrell, T., Shelhamer, E., Wang, D.: Back to the source: Diffusion-driven adaptation to test-time corruption. In: CVPR. pp. 11786--11796 (2023)

\bibitem{goodfellow2014generative}
Goodfellow, I., Pouget-Abadie, J., Mirza, M., Xu, B., Warde-Farley, D., Ozair, S., Courville, A., Bengio, Y.: Generative adversarial nets. In: NIPS. vol.~27 (2014)

\bibitem{gupta2016synthetic}
Gupta, A., Vedaldi, A., Zisserman, A.: Synthetic data for text localisation in natural images. In: CVPR. pp. 2315--2324 (2016)

\bibitem{ho2020denoising}
Ho, J., Jain, A., Abbeel, P.: Denoising diffusion probabilistic models. vol.~33, pp. 6840--6851 (2020)

\bibitem{ho2022cascaded}
Ho, J., Saharia, C., Chan, W., Fleet, D.J., Norouzi, M., Salimans, T.: Cascaded diffusion models for high fidelity image generation. JMLR  \textbf{23},  47--1 (2022)

\bibitem{ho2021classifierfree}
Ho, J., Salimans, T.: Classifier-free diffusion guidance. In: NIPS Workshop (2021)

\bibitem{hu2023gaia}
Hu, A., Russell, L., Yeo, H., Murez, Z., Fedoseev, G., Kendall, A., Shotton, J., Corrado, G.: Gaia-1: A generative world model for autonomous driving. arXiv preprint arXiv:2309.17080  (2023)

\bibitem{isola2017image}
Isola, P., Zhu, J.Y., Zhou, T., Efros, A.A.: Image-to-image translation with conditional adversarial networks. In: CVPR. pp. 1125--1134 (2017)

\bibitem{jaderberg2016synthetic}
Jaderberg, M., Simonyan, K., Vedaldi, A., Zisserman, A.: Synthetic data and artificial neural networks for natural scene text recognition. In: NIPS Workshop (2014)

\bibitem{7333942}
Karatzas, D., Gomez-Bigorda, L., Nicolaou, A., Ghosh, S., Bagdanov, A., Iwamura, M., Matas, J., Neumann, L., Chandrasekhar, V.R., Lu, S., Shafait, F., Uchida, S., Valveny, E.: Icdar competition on robust reading. In: ICDAR. pp. 1156--1160 (2015)

\bibitem{6628859}
Karatzas, D., Shafait, F., Uchida, S., Iwamura, M., Bigorda, L.G.i., Mestre, S.R., Mas, J., Mota, D.F., Almazàn, J.A., de~las Heras, L.P.: Icdar competition on robust reading. In: ICDAR. pp. 1484--1493 (2013)

\bibitem{karras2019style}
Karras, T., Laine, S., Aila, T.: A style-based generator architecture for generative adversarial networks. In: CVPR. pp. 4401--4410 (2019)

\bibitem{kim2017learning}
Kim, T., Cha, M., Kim, H., Lee, J.K., Kim, J.: Learning to discover cross-domain relations with generative adversarial networks. In: ICML. pp. 1857--1865 (2017)

\bibitem{kingma2014adam}
Kingma, D.P., Ba, J.: Adam: A method for stochastic optimization. arXiv preprint arXiv:1412.6980  (2014)

\bibitem{krasin2017openimages}
Krasin, I., Duerig, T., Alldrin, N., Ferrari, V., Abu-El-Haija, S., Kuznetsova, A., Rom, H., Uijlings, J., Popov, S., Veit, A., et~al.: Openimages: A public dataset for large-scale multi-label and multi-class image classification. Dataset available from https://github. com/openimages  \textbf{2}(3), ~18 (2017)

\bibitem{krishnan2023textstylebrush}
Krishnan, P., Kovvuri, R., Pang, G., Vassilev, B., Hassner, T.: Textstylebrush: Transfer of text aesthetics from a single example. PAMI  (2023)

\bibitem{krylov2021open}
Krylov, I., Nosov, S., Sovrasov, V.: Open images v5 text annotation and yet another mask text spotter. In: ACML. pp. 379--389 (2021)

\bibitem{kwon2023diffusionbased}
Kwon, G., Ye, J.C.: Diffusion-based image translation using disentangled style and content representation. In: ICLR (2023)

\bibitem{lee2021rewritenet}
Lee, J., Kim, Y., Kim, S., Yim, M., Shin, S., Lee, G., Park, S.: Rewritenet: Reliable scene text editing with implicit decomposition of text contents and styles. arXiv preprint arXiv:2107.11041  (2021)

\bibitem{li2021few}
Li, C., Taniguchi, Y., Lu, M., Konomi, S.: Few-shot font style transfer between different languages. In: WACV. pp. 433--442 (2021)

\bibitem{liu2022character}
Liu, R., Garrette, D., Saharia, C., Chan, W., Roberts, A., Narang, S., Blok, I., Mical, R., Norouzi, M., Constant, N.: Character-aware models improve visual text rendering. arXiv preprint arXiv:2212.10562  (2022)

\bibitem{liu2022xmp}
Liu, W., Liu, F., Ding, F., He, Q., Yi, Z.: Xmp-font: self-supervised cross-modality pre-training for few-shot font generation. In: CVPR. pp. 7905--7914 (2022)

\bibitem{lyu2017auto}
Lyu, P., Bai, X., Yao, C., Zhu, Z., Huang, T., Liu, W.: Auto-encoder guided gan for chinese calligraphy synthesis. In: ICDAR. vol.~1, pp. 1095--1100. IEEE (2017)

\bibitem{meng2022sdedit}
Meng, C., He, Y., Song, Y., Song, J., Wu, J., Zhu, J.Y., Ermon, S.: {SDE}dit: Guided image synthesis and editing with stochastic differential equations. In: ICLR (2022)

\bibitem{Mishra2009SceneTR}
Mishra, A., Karteek, A., Jawahar, C.V.: Scene text recognition using higher order language priors. In: BMVC (2009)

\bibitem{nakamura2017scene}
Nakamura, T., Zhu, A., Yanai, K., Uchida, S.: Scene text eraser. In: ICDAR. vol.~1, pp. 832--837. IEEE (2017)

\bibitem{nayef2019icdar2019}
Nayef, N., Patel, Y., Busta, M., Chowdhury, P.N., Karatzas, D., Khlif, W., Matas, J., Pal, U., Burie, J.C., Liu, C.l., et~al.: Icdar2019 robust reading challenge on multi-lingual scene text detection and recognition—rrc-mlt-2019. In: ICDAR. pp. 1582--1587 (2019)

\bibitem{pmlr-v139-nichol21a}
Nichol, A.Q., Dhariwal, P.: Improved denoising diffusion probabilistic models. In: ICML. vol.~139, pp. 8162--8171 (2021)

\bibitem{Noguchi_2024_WACV}
Noguchi, C., Fukuda, S., Yamanaka, M.: Scene text image super-resolution based on text-conditional diffusion models. In: WACV. pp. 1485--1495 (2024)

\bibitem{park2021multiple}
Park, S., Chun, S., Cha, J., Lee, B., Shim, H.: Multiple heads are better than one: Few-shot font generation with multiple localized experts. In: ICCV. pp. 13900--13909 (2021)

\bibitem{peng2023diffusion}
Peng, D., Hu, P., Ke, Q., Liu, J.: Diffusion-based image translation with label guidance for domain adaptive semantic segmentation. In: CVPR. pp. 808--820 (2023)

\bibitem{rombach2022high}
Rombach, R., Blattmann, A., Lorenz, D., Esser, P., Ommer, B.: High-resolution image synthesis with latent diffusion models. In: CVPR. pp. 10684--10695 (2022)

\bibitem{ronneberger2015u}
Ronneberger, O., Fischer, P., Brox, T.: U-net: Convolutional networks for biomedical image segmentation. In: MICCAI. pp. 234--241 (2015)

\bibitem{roy2020stefann}
Roy, P., Bhattacharya, S., Ghosh, S., Pal, U.: Stefann: scene text editor using font adaptive neural network. In: CVPR. pp. 13228--13237 (2020)

\bibitem{shen2023difftalk}
Shen, S., Zhao, W., Meng, Z., Li, W., Zhu, Z., Zhou, J., Lu, J.: Difftalk: Crafting diffusion models for generalized audio-driven portraits animation. In: CVPR. pp. 1982--1991 (2023)

\bibitem{7801919}
Shi, B., Bai, X., Yao, C.: An end-to-end trainable neural network for image-based sequence recognition and its application to scene text recognition. PAMI  \textbf{39}(11),  2298--2304 (2017)

\bibitem{shi2017icdar2017}
Shi, B., Yao, C., Liao, M., Yang, M., Xu, P., Cui, L., Belongie, S., Lu, S., Bai, X.: Icdar2017 competition on reading chinese text in the wild (rctw-17). In: ICDAR. vol.~1, pp. 1429--1434 (2017)

\bibitem{singh2021textocr}
Singh, A., Pang, G., Toh, M., Huang, J., Galuba, W., Hassner, T.: Textocr: Towards large-scale end-to-end reasoning for arbitrary-shaped scene text. In: CVPR. pp. 8802--8812 (2021)

\bibitem{pmlr-v37-sohl-dickstein15}
Sohl-Dickstein, J., Weiss, E., Maheswaranathan, N., Ganguli, S.: Deep unsupervised learning using nonequilibrium thermodynamics. In: ICML. vol.~37, pp. 2256--2265 (2015)

\bibitem{sun2019icdar}
Sun, Y., Ni, Z., Chng, C.K., Liu, Y., Luo, C., Ng, C.C., Han, J., Ding, E., Liu, J., Karatzas, D., et~al.: Icdar 2019 competition on large-scale street view text with partial labeling-rrc-lsvt. In: ICDAR. pp. 1557--1562 (2019)

\bibitem{tang2022few}
Tang, L., Cai, Y., Liu, J., Hong, Z., Gong, M., Fan, M., Han, J., Liu, J., Ding, E., Wang, J.: Few-shot font generation by learning fine-grained local styles. In: CVPR. pp. 7895--7904 (2022)

\bibitem{tuo2023anytext}
Tuo, Y., Xiang, W., He, J.Y., Geng, Y., Xie, X.: Anytext: Multilingual visual text generation and editing (2024)

\bibitem{wang2023cf}
Wang, C., Zhou, M., Ge, T., Jiang, Y., Bao, H., Xu, W.: Cf-font: Content fusion for few-shot font generation. In: CVPR. pp. 1858--1867 (2023)

\bibitem{6126402}
Wang, K., Babenko, B., Belongie, S.: End-to-end scene text recognition. In: ICCV. pp. 1457--1464 (2011)

\bibitem{wang2018high}
Wang, T.C., Liu, M.Y., Zhu, J.Y., Tao, A., Kautz, J., Catanzaro, B.: High-resolution image synthesis and semantic manipulation with conditional gans. In: CVPR. pp. 8798--8807 (2018)

\bibitem{wang2021real}
Wang, X., Xie, L., Dong, C., Shan, Y.: Real-esrgan: Training real-world blind super-resolution with pure synthetic data. In: ICCV. pp. 1905--1914 (2021)

\bibitem{wang2023stylediffusion}
Wang, Z., Zhao, L., Xing, W.: Stylediffusion: Controllable disentangled style transfer via diffusion models. In: ICCV. pp. 7677--7689 (2023)

\bibitem{10.1145/3343031.3350929}
Wu, L., Zhang, C., Liu, J., Han, J., Liu, J., Ding, E., Bai, X.: Editing text in the wild. In: ACM MM. p. 1500–1508 (2019)

\bibitem{xie2021dg}
Xie, Y., Chen, X., Sun, L., Lu, Y.: Dg-font: Deformable generative networks for unsupervised font generation. In: CVPR. pp. 5130--5140 (2021)

\bibitem{yang2020swaptext}
Yang, Q., Huang, J., Lin, W.: Swaptext: Image based texts transfer in scenes. In: CVPR. pp. 14700--14709 (2020)

\bibitem{yang2019tet}
Yang, S., Liu, J., Wang, W., Guo, Z.: Tet-gan: Text effects transfer via stylization and destylization. In: AAAI. vol.~33, pp. 1238--1245 (2019)

\bibitem{yang2024glyphcontrol}
Yang, Y., Gui, D., Yuan, Y., Liang, W., Ding, H., Hu, H., Chen, K.: Glyphcontrol: Glyph conditional control for visual text generation. In: NIPS. vol.~36 (2024)

\bibitem{yi2017dualgan}
Yi, Z., Zhang, H., Tan, P., Gong, M.: Dualgan: Unsupervised dual learning for image-to-image translation. In: ICCV. pp. 2849--2857 (2017)

\bibitem{yim2021synthtiger}
Yim, M., Kim, Y., Cho, H.C., Park, S.: Synthtiger: Synthetic text image generator towards better text recognition models. In: ICDAR. pp. 109--124. Springer (2021)

\bibitem{zhan2018verisimilar}
Zhan, F., Lu, S., Xue, C.: Verisimilar image synthesis for accurate detection and recognition of texts in scenes. In: ECCV. pp. 249--266 (2018)

\bibitem{zhan2019ga}
Zhan, F., Xue, C., Lu, S.: Ga-dan: Geometry-aware domain adaptation network for scene text detection and recognition. In: CVPR. pp. 9105--9115 (2019)

\bibitem{zhan2019spatial}
Zhan, F., Zhu, H., Lu, S.: Spatial fusion gan for image synthesis. In: CVPR. pp. 3653--3662 (2019)

\bibitem{zhang2019icdar}
Zhang, R., Zhou, Y., Jiang, Q., Song, Q., Li, N., Zhou, K., Wang, L., Wang, D., Liao, M., Yang, M., et~al.: Icdar 2019 robust reading challenge on reading chinese text on signboard. In: ICDAR. pp. 1577--1581 (2019)

\bibitem{zhang2019ensnet}
Zhang, S., Liu, Y., Jin, L., Huang, Y., Lai, S.: Ensnet: Ensconce text in the wild. In: AAAI. vol.~33, pp. 801--808 (2019)

\bibitem{zhang2018separating}
Zhang, Y., Zhang, Y., Cai, W.: Separating style and content for generalized style transfer. In: CVPR. vol.~1 (2018)

\bibitem{zhang2017uber}
Zhang, Y., Gueguen, L., Zharkov, I., Zhang, P., Seifert, K., Kadlec, B.: Uber-text: A large-scale dataset for optical character recognition from street-level imagery. In: CVPR Workshops. vol.~2017, p.~5 (2017)

\bibitem{zhang2023inversion}
Zhang, Y., Huang, N., Tang, F., Huang, H., Ma, C., Dong, W., Xu, C.: Inversion-based style transfer with diffusion models. In: CVPR. pp. 10146--10156 (2023)

\bibitem{zhu2017unpaired}
Zhu, J.Y., Park, T., Isola, P., Efros, A.A.: Unpaired image-to-image translation using cycle-consistent adversarial networks. In: ICCV. pp. 2223--2232 (2017)

\bibitem{zhu2023conditional}
Zhu, Y., Li, Z., Wang, T., He, M., Yao, C.: Conditional text image generation with diffusion models. In: CVPR. pp. 14235--14245 (2023)

\end{thebibliography}

\clearpage

\appendix

\section{Dataset Details}

We selected English for the source language due to the widespread availability of datasets.
As described in the main text, we incorporated 12 publicly available real datasets, culminating in a total of 2.56M text images aling with their corresponding text labels. 

Regarding the target languages, we selected five languages: Arabic, Bengali, Chinese, Japanese, and Korean.  To produce synthetic and plain text images for these five languages, as well as English, we utilized SynthTIGER \cite{yim2021synthtiger}, producing 2M images per language. For producing these synthetic text images, we utilized SynthTIGER's default settings, including background image color, texture, text layouts, text styles, midground text,  geometric transformation, postprocessing.

The word set for each language was derived from the EasyOCR repository \cite{easyocr} and Wikipedia pages using an API. These word sets consisted of 36 Arabic, 74 Bengali, 6,614 Chinese, 2,100 Japanese, and 1,471 Korean unique characters, respectively.
We randomly choose one word from the word set to create the corresponding synthetic and plain text images.
We sourced free font files primarily from Google Fonts. However, due to a scarcity of Chinese and Japanese font files in Google Fonts, additional collections were made from various online sources. The final count of font files obtained for each language was 64 for Arabic, 20 for Bengali, 24 for Korean, 33 for Chinese, and 98 for Japanese.

\begin{table}[t!]
  \centering
    \begin{tabular}{c|c}
        \hline
       Diffusion Steps (for Training) & 1000\\
       Diffusion Steps (for EDB Guidance) & 100\\
       Diffusion Steps (for FE Guidance) & 200\\
       Noies Schedule & cosine \\
       Channels & 192  \\
       Channel Multiplier & 1,1,2,2,4,4 \\
       Number of Heads & 3 \\
       Number of ResBlocks & 3 \\
       Batch Size & 128  \\
       Learning Rate & 1e-4  \\
       Dropout & 0.1  \\
       Iterations & 1.5M  \\
       Embedding Dimension & 768 \\
       Attention Resolution & 32,16,8  \\
      \hline
      \end{tabular}
\caption{Hyperparameters for the diffusion model in our proposed framework.}
\label{table:hyperparameters}
\end{table}

\section{Implementation Details}

The implementation of the diffusion model in our proposed framework was based on the code released by the authors in \cite{pmlr-v139-nichol21a,NEURIPS2021_49ad23d1}. 
For the training process, we employed the AdamW \cite{kingma2014adam} optimizer with a learning rate $1\times 10^{-4}$. Our diffusion model underwent training for 1.5M iterations using a batch size of 128. During training, the diffusion time step $T$ was set to 1000. At inference, $T$ was set to 100 for the FDB Guidance and 200 for the FE Guidance. 
The hyperparameters of the model architecture are presented in Tab. \ref{table:hyperparameters}.

We selected $w_{min}=1$, $w_{max}=6$ as the hyperparameters for FDB Guidance across all languages. For the hyperparameters of FE Guidance, we selected $n=120$. In addition, for $k_{1}$ and $k_{2}$, we set $(k_1, k_2)=(2,1)$ for Arabic, $(3,1)$ for Bengali, $(1,6)$ for Chinese, $(1,6)$ for Japanese, and $(1,1)$ for Korean. During the inference phase with the FE Guidance, we also utilized the FDB Guidance to schedule the guidance scales.
All models utilized the same hyperparameters, with the exception of those related to the two guidance techniques.
All images, both for training and inference, were resized to a resolution of $32\times 128$.

As explained in the main text, the diffusion model in our framework is conditioned on two types of inputs. The first is a plain image $y$, and the second is a binary variable $c$ with two states:  
\textit{synth} and \textit{real}.
To condition the diffusion model on a plain image $y$, the $t$th input image $x_t$ is replaced by $[x_t,y]$, where $[\cdot,\cdot]$ stands for the operation of concatenation in the channel dimension.
For the binary variable, we first represented it as a one-hot vector. This vector is then embedded to match the dimension of the time-embedding vectors. Subsequently, the diffusion model received this embedded one-hot vector in two ways: by adding it to the time-embedding vector, and through the use of cross-attention modules.

The diffusion model in our proposed framework adopts classifier-free guidance \cite{ho2021classifierfree} to generate text images that more accurately reflect the textual content of the input plain text images.
Utilizing classifier-free guidance necessitates the use of an unconditional denoising model $\epsilon_\theta (x_t, t)$ during the inference phase.
Following prior studies, this unconditional model is acquired through the joint training of both unconditional model $\epsilon_\theta (x_t, t)$ and conditional model $\epsilon_\theta (x_t, t, c, y)$.
Specifically, the conditions $c$ and $y$ were probabilistically omitted at a consistent rate during the training. In our framework, these conditions were omitted with a 10\% probability. For omitting condition $c$, we used all-zero vectors, and for omitting condition $y$, we used completely white images.

\section{ Qualitative Evaluation for the Effectiveness of FE Guidance}

\begin{figure*}[h!]
    \begin{tabular}{c}
      \begin{minipage}[t]{0.95\linewidth}
        \centering
        \includegraphics[width=12.0cm]{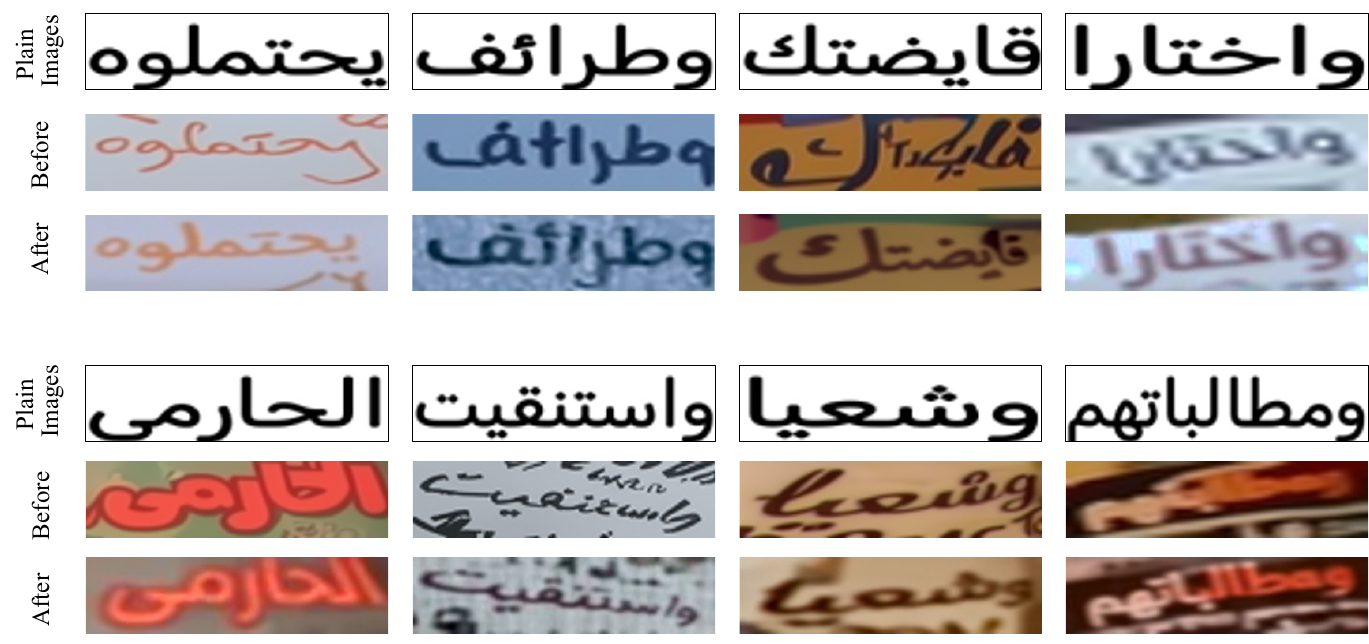}
      \end{minipage}
      
    \end{tabular}
    \caption{Arabic text images before and after applying FE Guidance. The top row shows plain text images, the middle row displays examples before FE Guidance is applied, and the bottom row demonstrates examples after the application of FE Guidance.}
\label{fig:supp_ar}
\end{figure*}

\begin{figure*}[t!]
    \begin{tabular}{c}
      \begin{minipage}[t]{0.95\linewidth}
        \centering
        \includegraphics[width=12.0cm]{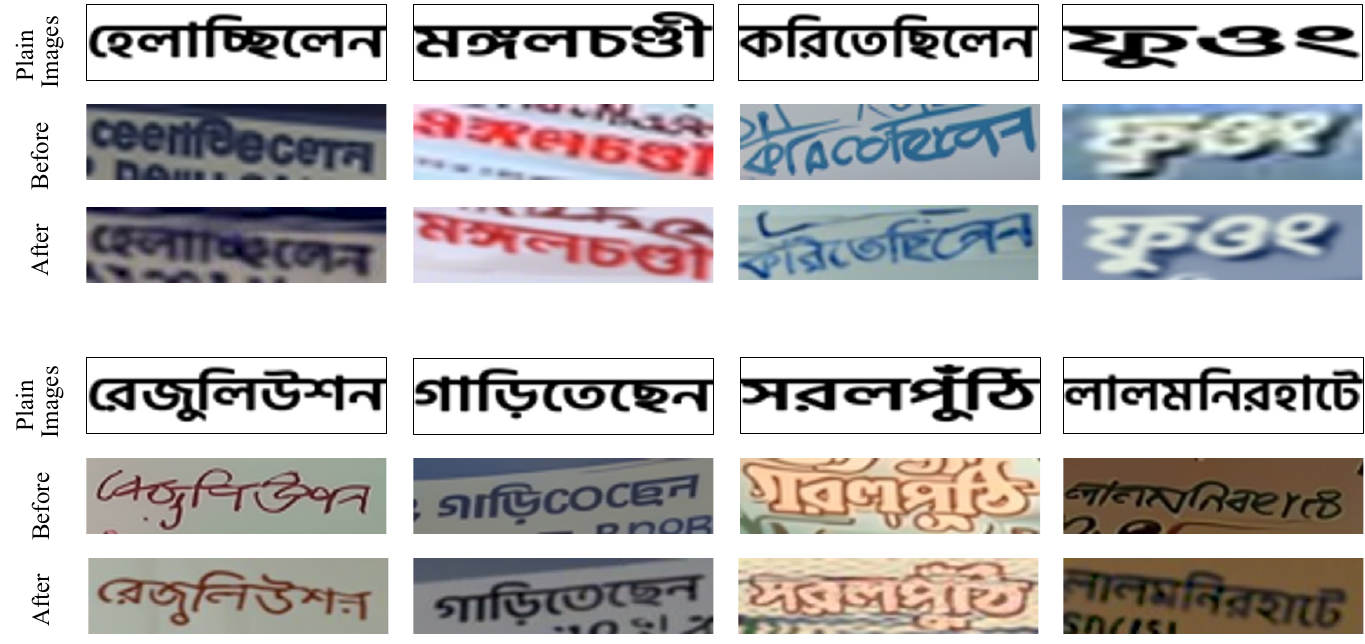}
      \end{minipage}
      
    \end{tabular}
    \caption{Bengali text images before and after applying FE Guidance. The top row shows plain text images, the middle row displays examples before FE Guidance is applied, and the bottom row demonstrates examples after the application of FE Guidance.}
\label{fig:supp_bn}
\end{figure*}

\begin{figure*}[t!]
    \begin{tabular}{c}
      \begin{minipage}[t]{0.95\linewidth}
        \centering
        \includegraphics[width=12.0cm]{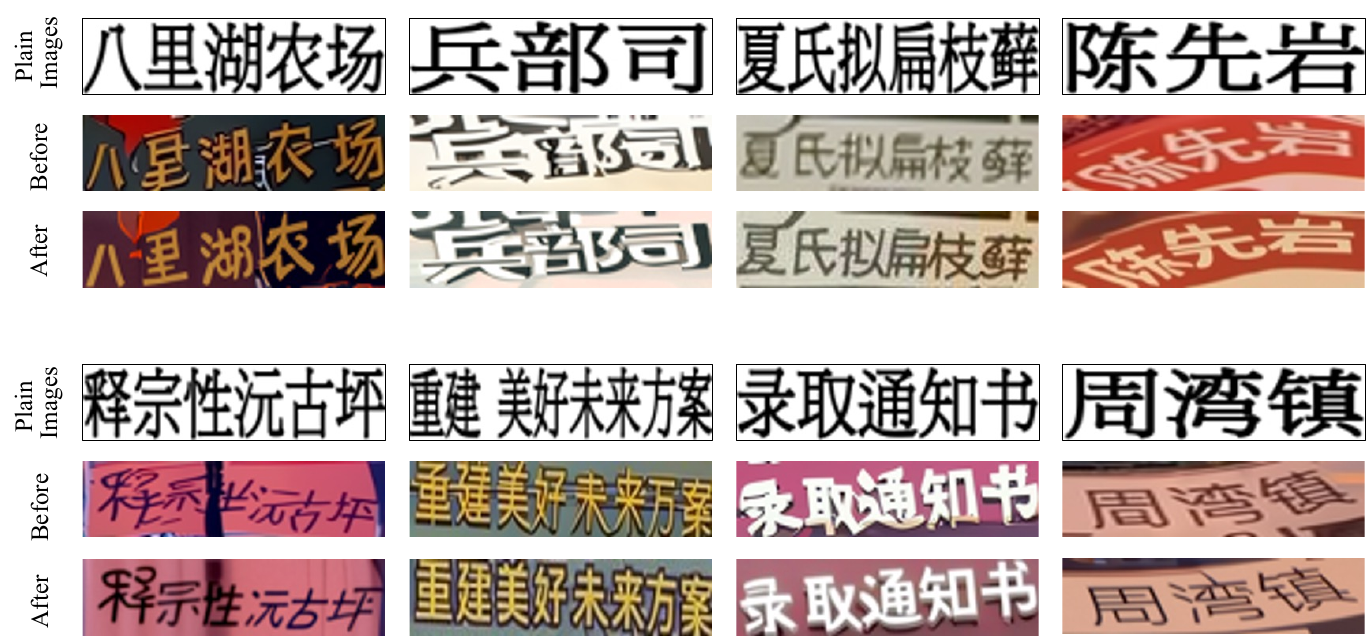}
      \end{minipage}
      
    \end{tabular}
    \caption{Chinese text images before and after applying FE Guidance. The top row shows plain text images, the middle row displays examples before FE Guidance is applied, and the bottom row demonstrates examples after the application of FE Guidance.}
\label{fig:supp_ch}
\end{figure*}

\begin{figure*}[t!]
    \begin{tabular}{c}
      \begin{minipage}[t]{0.95\linewidth}
        \centering
        \includegraphics[width=12.0cm]{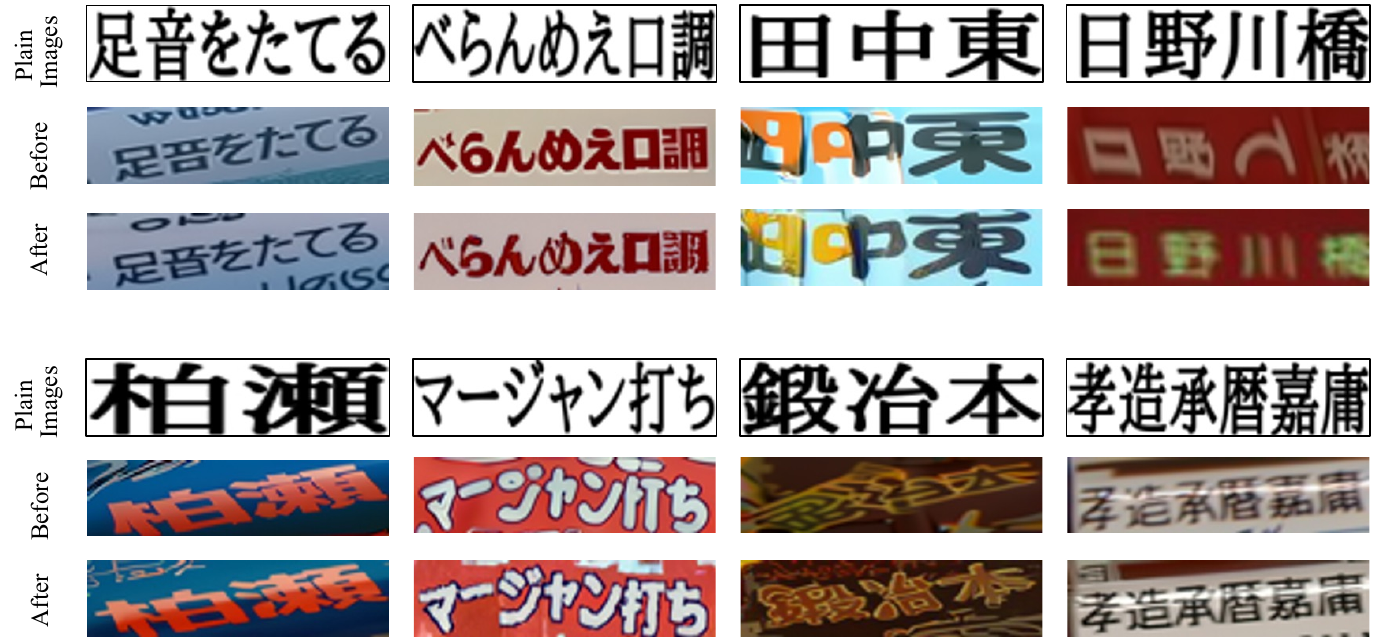}
      \end{minipage}
      
    \end{tabular}
    \caption{Japanese text images before and after applying FE Guidance. The top row shows plain text images, the middle row displays examples before FE Guidance is applied, and the bottom row demonstrates examples after the application of FE Guidance.}
\label{fig:supp_ja}
\end{figure*}

\begin{figure*}[t!]
    \begin{tabular}{c}
      \begin{minipage}[t]{0.95\linewidth}
        \centering
        \includegraphics[width=12.0cm]{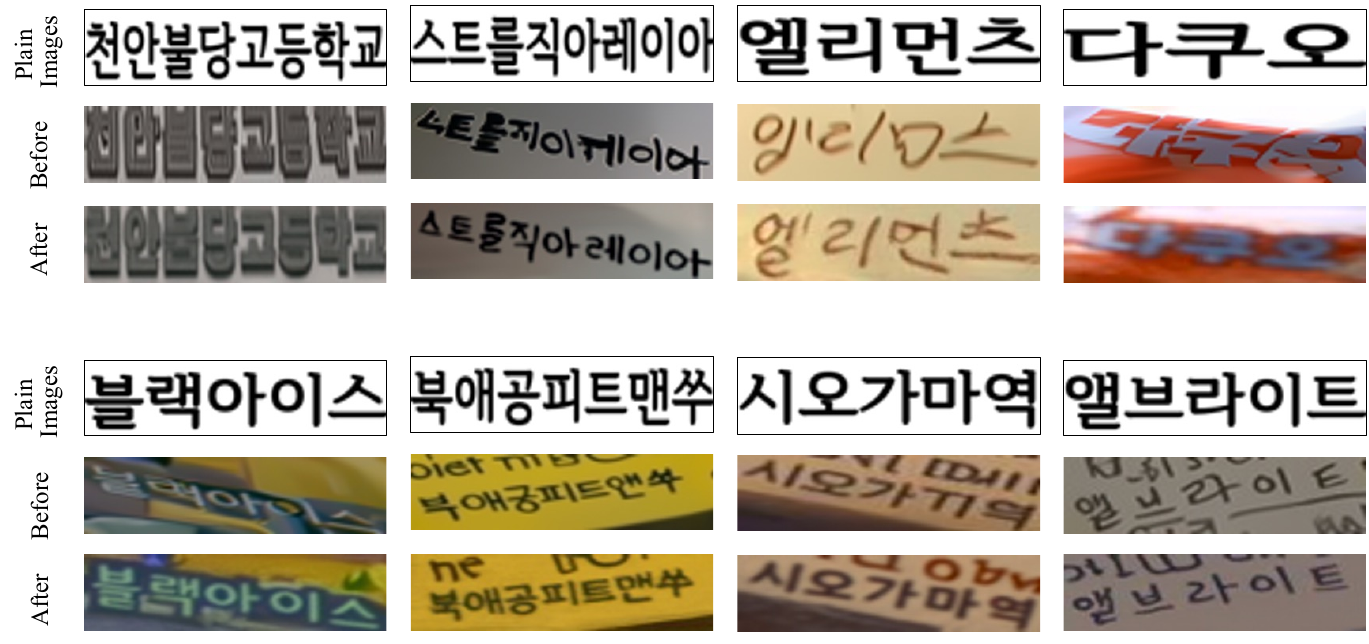}
      \end{minipage}
      
    \end{tabular}
    \caption{Korean text images before and after applying FE Guidance. The top row shows plain text images, the middle row displays examples before FE Guidance is applied, and the bottom row demonstrates examples after the application of FE Guidance.}
\label{fig:supp_ko}
\end{figure*}

In Figs. \ref{fig:supp_ar}-\ref{fig:supp_ko}, we showcase examples to qualitatively evaluate the efficacy of FE Guidance. These figures display samples of Arabic, Bengali, Chinese, Japanese, and Korean text images, in that order. The first row depicts plain text images, the second row presents examples prior to the application of FE Guidance, and the third row illustrates examples following the application of FE Guidance.

The efficacy of the FE Guidance in correcting the textual content of generated images improves as the value of $k_1$ increases and the value of $k_2$ decreases. As described in the main text, we set higher $k_1$ values and smaller $k_2$ values for Arabic and Bengali languages. The Arabic and Bengali text images shown in Figs. \ref{fig:supp_ar} and \ref{fig:supp_bn} reveal that, even through the textual contents of the text images before applying FE Guidance significantly deviate from the intended output, the FE Guidance is capable of successfully rectifying it. 
Nonetheless, as explained in the main text, this effectiveness is achieved at the expense of the style of real text images. Indeed, after applying the FE Guidance, the styles of these text images tend to resemble those of synthetic text images more closely.
Conversely, for Chinese and Japanese, where we used lower $k_1$ and higher $k_2$ values, the rectification capability is not as pronounced as for Arabic and Bengali. 
Our experiments indicate that our framework, without FE Guidance, generates more accurate text images for Chinese and Japanese than for Arabic and Bengali. Thus, in many instances, the lower $k_1$ and higher $k_2$ settings are sufficient. The benefit of using lower $k_1$ and higher $k_2$ values are that our framework can create text images while maintaining the style of real text images.

\end{document}